# The Role of Orthographic Consistency in Multilingual Embedding Models for Text Classification in Arabic-Script Languages

Abdulhady Abas Abdullah[1,2], Amir H. Gandomi[3], Tarik A Rashid[1], Seyedali Mirjalili[4], Laith Abualigah[5], Milena Živković[6], Hadi Veisi[7],

[1]Artificial Intelligence and Innovation Centre, University of Kurdistan Hewler

[2]Computer Science Department, Faculty of Science, Soran University, Soran, Kurdistan, Iraq

[3]Faculty of Engineering and Information Technology, University of Technology Sydney, Ultimo, Sydney, 2007, NSW, Australia

[4]Center for Artificial Intelligence Research and optimisation, Torrens University Australia, Brisbane, QLD 4006, Australia

[5]School of Engineering and Technology, Sunway University Malaysia, Petaling Jaya 27500, Malaysia

[6]Department of physics, University of Kragujevac, Faculty of science, Kragujevac, Serbia

[7]College of Interdisciplinary Science and Technologies, University of Tehran, Tehran, Iran

## Abstract

In the dynamic field of natural language processing, the advent of models such as mBERT and XLM-RoBERTa has heralded an era of unprecedented multilingual capability. These models promise to address dozens of languages with a single architecture, yet in practice, they often falter when confronted with languages that share a common script but differ in orthographic conventions and cultural contexts. Among the languages using the Arabic script, Kurdish Sorani, Arabic, Persian, and Urdu, this limitation is particularly acute, leaving room for more specialized solutions. Motivated by this challenge, we set out to design models that truly understand the script they encounter. We introduced the Arabic Script (AS) RoBERTa family, comprising four distinct RoBERTa-based models, each pre-trained on a vast corpus of Arabic-script text curated for its target language. By tailoring the pre-training phase to the unique statistical and orthographic properties of Kurdish Sorani, Arabic, Persian, and Urdu, we sought to capture the subtle patterns that generic multilingual models often overlook. When fine-tuned on language-specific classification tasks, each AS-RoBERTa variant achieved accuracy gains of 2 to 5 percentage points over the XLM-RoBERTa and mBERT baselines. An ablation study confirmed that script-focused pre-training was the cornerstone of these improvements. A detailed error analysis, visualized through confusion matrices, further revealed how shared script features and domain-specific cultural content shaped each model's performance, illuminating both strengths and remaining challenges. Embracing script-centric specialization yields tangible benefits for text classification in languages that share the Arabic script. This work underscores the value of language-specific pre-training and invites future research to explore script-based strategies across diverse linguistic landscapes. Through this journey, we demonstrate that deep understanding arises not from universalism alone but from respect for the unique characteristics that each language and script contribute to the task.

## Introduction

Text classification is a fundamental task in natural language processing (NLP) with applications from sentiment analysis to topic categorization. Recent advances in transformer-based language models have greatly improved text classification, especially with the advent of large pre-trained models like BERT and its multilingual variants. Multilingual BERT (mBERT), trained on 104 languages, and XLM-RoBERTa (XLM-R), trained on 100 languages, have demonstrated surprising cross-lingual generalization ability. (Nabiilah et al., 2024, Fan et al., 2021, Wiciaputra et al., 2021). These models can be fine-tuned once and applied to multiple languages, which is appealing for low-resource scenarios. However, studies have found that for a given language, a dedicated monolingual model often achieves higher accuracy than a multilingual model on the same task. (Oladipo et al., 2023). This gap is especially evident in languages with abundant data, where a language-specific model can fully exploit the nuances of that language's vocabulary and grammar. (Fuad and Al-Yahya, 2022).

Arabic-script languages present an interesting case study for multilingual classification. Languages such as Arabic, Persian (Farsi), Urdu, and the Sorani dialect of Kurdish all use Perso-Arabic scripts, sharing a writing system but belonging to different language families (Semitic, Indo-European, Indo-Aryan) with largely distinct vocabularies (Abudalfa et al., 2024, Mekki et al., 2022). Prior works have developed BERT-based models for Arabic, i.e., AraBERT (El-Alami et al., 2022) and Persian, i.e., ParsBERT (Farahani et al., 2021), which achieved state-of-the-art results on native benchmarks by outscoring multilingual baselines. In contrast, Kurdish and Urdu have remained relatively under-served in this regard. Kurdish, in particular, is considered a low-resource language with limited NLP resources and datasets. Until recently, most Kurdish text classification efforts relied on traditional machine learning or translated models, and the language still "suffers from fundamental issues and challenges in NLP". Urdu has more digital presence, but dedicated Urdu-language transformer models are still in early development; researchers have begun to compile large Urdu news corpora (e.g., 50k+ articles) to facilitate training of such models (Awlla et al., 2025, Malik et al., 2023). This paper addresses these gaps by building language-specific RoBERTa models for Kurdish, Arabic, Persian, and Urdu and evaluating their effectiveness on text classification. Figure 1 demonstration mult corpus structure.

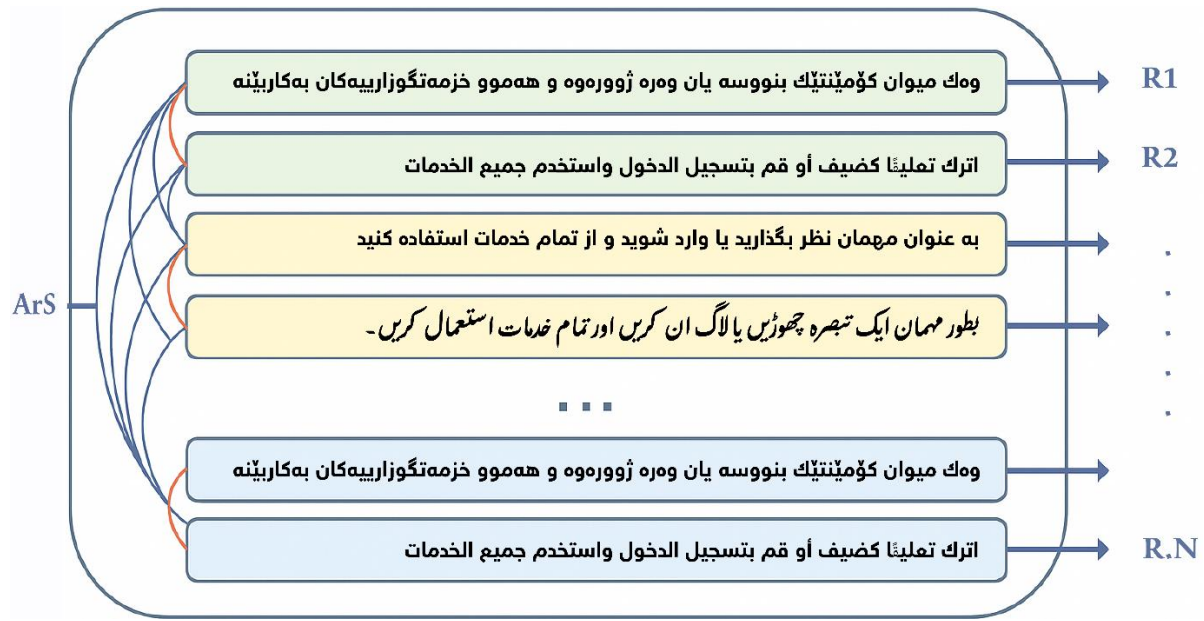

**Figure 1.** Multilingual Arabic-Script Corpora Retrieval.

It is a script-centric multilingual pre-training strategy that can yield superior classification performance across Arabic-script languages when compared with a broadly trained multilingual model. While shared orthography facilitates transfer of character-level representations, significant lexical, morphological, and syntactic divergences among Sorani Kurdish, Arabic, Persian, and Urdu may dilute language-specific cues within a wholly general architecture. To address this, we pre-trained a single RoBERTa model on a comprehensive Arabic-script corpus encompassing text from all four target languages to capture both shared and language-unique patterns in one unified framework. We then fine-tuned this AS-RoBERTa model on language-tailored classification benchmarks, including news-topic categorization and sentiment analysis, to quantify the gains afforded by script-focused, multilingual pre-training.

The key contributions of this work can be summarized as follows. First, we introduce AS-RoBERTa, a RoBERTa-based transformer model pre-trained on a large unified Arabic-script corpus that integrates text from Sorani Kurdish, Arabic, Persian, and Urdu, thereby capturing both shared orthographic features and language-specific patterns within a single multilingual framework. Second, through systematic fine-tuning on language-tailored benchmarks, including news-topic categorization and sentiment analysis, we demonstrate that AS-RoBERTa consistently outperforms leading multilingual baselines (XLM-RoBERTa and mBERT), yielding absolute accuracy gains of 2 to 5 percent across all four target languages. Third, an ablation study isolates the impact of script-focused pre-training, confirming that the inclusion of Arabic-script text during pre-training is the primary driver of these performance improvements. Finally, we provide an in-depth qualitative error analysis using confusion matrices to reveal how script similarity and cultural domain differences influence classification outcomes, and we discuss the practical implications of these findings for the development of effective cross-lingual NLP systems in Arabic-script contexts.

The remainder of this paper is organized as follows. Section II basic concept works on multilingual and monolingual transformer models and prior NLP research for each of the four languages. Section III details our methodology, including the RoBERTa model architecture and the training pipeline. Section IV describes the experimental setup, datasets, and training configurations. Section V presents the results, comparing performance across models and languages. Section VI provides a discussion on error analysis and implications of our findings. Finally, Section VII concludes the paper and outlines future directions.

## Basic Concept

The development of multilingual BERT (mBERT) and subsequent models like XLM-RoBERTa (Conneau et al., 2020) has significantly advanced cross-lingual NLP capabilities. These models, trained on massive multilingual corpora (e.g., XLM-R's 2.5TB across 100 languages), establish shared embedding spaces that enable effective zero-shot transfer. However, research by Wu & Dredze (2020) reveals important limitations: while multilingual models perform well on high-resource languages, their effectiveness diminishes for low-resource languages, though they still outperform naive monolingual models trained on limited data. This suggests multilingual training provides a beneficial regularization effect. Recent work has explored adaptation techniques like vocabulary augmentation and continued pre-training to enhance performance on specific languages, as well as zero-shot approaches using models like XLM-RoBERTa [??]. While zero-shot methods offer practical advantages for rapid deployment, as demonstrated by Manias et al. (2023) work on multilingual social media classification. They consistently underperform compared to properly fine-tuned models, reinforcing the value of language-specific adaptation when training data is available. Recent large-scale multilingual models such as mLongT5 (Uthus et al., 2023) and XGLM (Lin et al., 2022) have further advanced zero-shot transfer, demonstrating that parameter-efficient fine-tuning and prompt-based methods can yield substantial gains even in truly low-resource settings.

For many languages, dedicated monolingual BERT and RoBERTa models have demonstrated superior performance compared to multilingual counterparts. In Arabic, AraBERT (Antoun et al., 2021) achieved state-of-the-art results across various tasks by leveraging approximately 70 million words of pre-training data, significantly outperforming mBERT. Beyond AraBERT, Abdul-Mageed et al. (2021) introduced MARBERT, trained on 1 billion Arabic tweets plus diverse text, which excels particularly on dialectal Arabic. These models outperform multilingual baselines by several points on tasks like sentiment (e.g., Arabic tweets sentiment classification) and even approach human performance on certain benchmarks. The reason is that Arabic's complex morphology (e.g., templatic root-pattern structures, infixation, and clitics like the definite article and pronoun suffixes) is handled much better with an Arabic-specific vocabulary and pre-training. Multilingual models often split or misrepresent Arabic words (which are often agglutinative), whereas AraBERT/MARBERT learned whole-word representations for such forms. Similarly, ParsBERT (Farahani et al., 2021) established new benchmarks for Persian NLP through comprehensive pre-training on 3.9 billion words, showing consistent 2-5% improvements over multilingual baselines. These successes highlight how focused capacity on a single language, particularly those with rich morphological systems like Arabic and Persian, can yield substantial gains when sufficient pre-training data exists. However, similar efforts for other Arabic-script languages remain limited. In stark contrast to Arabic and Persian, Urdu (which uses a Persian-Arabic script and shares some vocabulary with Persian) has been *under-resourced* and until now lacked a comparable monolingual Transformer. For Urdu, researchers have primarily worked with fine-tuned mBERT variants or smaller BERT models, while Javed et al.'s work (2021) LSTM-based models on Urdu news classification demonstrated the potential for transformer-based approaches, highlighting the need for dedicated Urdu language models. Moreover, lightweight adaptation techniques, such as adapter layers (Houlsby et al., 2019), LoRA (Hu et al., 2021), and vocabulary extension strategies, have shown promising improvements on mid-sized corpora without requiring full model retraining.  Most prior Urdu NLP efforts have relied either on multilingual models or on classical approaches. For instance, Khan et al. (2022) compiled a new Urdu sentiment analysis dataset (9,312 annotated reviews across multiple domains) and fine-tuned mBERT on it. Their fine-tuned mBERT achieved an F1 score of 81.5%, substantially outperforming traditional machine learning classifiers (SVM, Naïve Bayes, etc.) on the same data. This demonstrates the power of Transformer representations even in a zero-shot context mBERT, despite not being Urdu-specific, provided a strong baseline for Urdu sentiment. However, 81.5% F1 still leaves considerable headroom, suggesting that a dedicated Urdu model might push this further.

Similarly, Kurdish (Sorani dialect) presents a particularly challenging case as a low-resource language with limited NLP tools and annotated datasets. Early approaches to Kurdish text classification relied on traditional machine learning methods, with sentiment analysis studies achieving only 66% accuracy using Naïve Bayes classifiers (Ahmadi and Bouallegue, 2015). The introduction of transformer models brought significant improvements, as shown by Badawi (2023), where multilingual BERT achieved 92% accuracy on Kurdish sentiment analysis, outperforming classical methods by a considerable margin. However, the absence of a dedicated Kurdish BERT or RoBERTa (Abdullah et al., 2024) The model has forced researchers to depend on multilingual alternatives. The Kurdish Sorani writing system, using a modified Arabic script with frequent inconsistencies (including Arabic letter substitutions), makes robust text normalization essential. Tools like the KLPT toolkit (Ahmadi, 2020) provide crucial normalization and tokenization support for Kurdish text processing. This gap is addressed by developing the first dedicated RoBERTa model for Kurdish, trained on an extensive corpus of approximately 188 million tokens. Complementary approaches like prompt-based fine-tuning (Liu et al., 2021) and contrastive pre-training (Gao et al., 2021) have recently been applied to low-resource languages, suggesting additional pathways to boost performance when labeled data remain scarce.

The literature establishes that language-specific transformers (e.g., AraBERT, ParsBERT) consistently outperform multilingual models when sufficient training data exists, while multilingual approaches (e.g., mBERT, XLM-R) remain stop-gap solutions for low-resource languages like Kurdish and Urdu. Our work advances this paradigm by (1) introducing the first dedicated RoBERTa models for Kurdish and Urdu, (2) providing curated datasets/preprocessing tools, and (3) offering the first systematic comparison of monolingual versus multilingual approaches across four Arabic-script languages, revealing new insights into script similarity effects. Though emergent multilingual systems (e.g., BLOOMZ (Yong et al., 2023)) show promise, our results reaffirm the value of language-specific models, particularly for morphologically rich languages.

## Proposed Method

In this section, we provide an extended, in-depth description of our pipeline (Figure 2), elaborating on the theoretical motivations, the dual-stage training objectives, the orthographic-aware masking mechanism, and the domain-adaptive, multi-tokenizer framework, before concluding with the design of our language-specific classification heads.

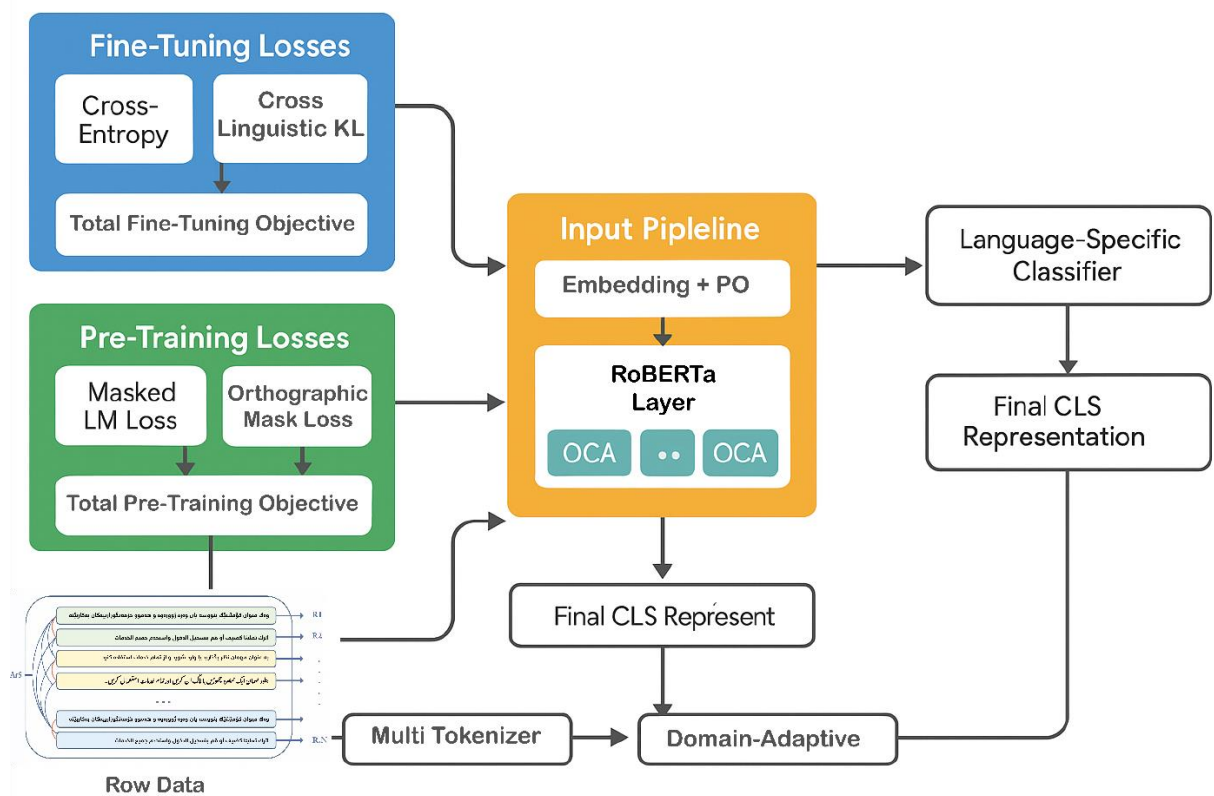

Figure 2. Language-routed classification system using specialized RoBERTa models for Arabic, Persian, Urdu and Kurdish.

### 3.1 Model Architecture

We begin by defining our input representation. Let $x = [x_1, \ldots, x_n]$ be the sequence of subword tokens obtained from an Arabic-script sentence. To capture complementary segmentation granularities, we employ two tokenizers-Byte-Pair Encoding (BPE) and WordPiece (WP)-and fuse their embeddings by averaging (Eq. 1).

$$H^{(0)} = \frac{1}{2}(\text{Emb}_{\text{BPE}}(x) + \text{Emb}_{\text{WP}}(x)) \qquad (1)$$

This fusion allows the model to leverage both the course, frequent-unit coverage of BPE and the finer, context-sensitive splits of WordPiece. By averaging, we maintain a simple yet effective mechanism for combining subword vocabularies without inflating model size.

**Algorithm 1:** Orthographic Consistency Pre-training & Domain-Adaptive Fine-Tuning

**Require:** Unlabeled corpus $D_n$; Labeled set $D_l = \{(x, y, \ell)\}$; backbone params $\theta_{\text{backbone}}$; Embeddings $\text{Emb}_{\text{BPE}}, \text{Emb}_{\text{WP}}$;
  #layers $L$, proj. size $m$, weights $\beta, \gamma$, lrs $\eta_{\text{pre}}, \eta_{\text{fine}}$

**Ensure :** Fine-tuned $\theta_{\text{backbone}}$, proj. $\theta_{\text{DA}}$, heads $\{W_\ell, b_\ell\}$

**Function** FuseEmbeddings($x$)
  **return** $\frac{1}{2}(\text{Emb}_{\text{BPE}}(x) + \text{Emb}_{\text{WP}}(x))$  // Eq. (1)

**Function** ForwardWithOCA($H^{(0)}$)
  $H \leftarrow H^{(0)}$
  **for** $i = 1$ **to** $L$ **do**
    $H \leftarrow \text{TransformerLayer}_i(H)$
    $H \leftarrow \text{OrthographicConsistencyAdapter}(H)$
  **return** $H$

**Function** Pretrain($D_n, \beta$)
  **for** *epoch* = 1 **to** $N_{\text{pre}}$ **do**
    **foreach** *batch* $x \in D_n$ **do**
      $M \leftarrow \text{random\_mask}(x)$
      $O \leftarrow \text{orth\_variants}(x)$
      $H^{(0)} \leftarrow \text{FuseEmbeddings}(x)$
      $H \leftarrow \text{ForwardWithOCA}(H^{(0)})$
      $L_{\text{MLM}} \leftarrow \text{MLM\_loss}(H, M)$  // Eq. (2)
      $L_{\text{orth}} \leftarrow \text{orth\_loss}(H, O)$  // Eq. (3)
      $L_{\text{pre}} \leftarrow L_{\text{MLM}} + \beta L_{\text{orth}}$
       // Eq. (4)
      $\theta_{\text{backbone}} \leftarrow \theta_{\text{backbone}} - \eta_{\text{pre}} \nabla L_{\text{pre}}$
  **return** $\theta_{\text{backbone}}$

**Function** Finetune($D_l, \gamma$)
  initialize $\theta_{\text{DA}}, \{W_\ell, b_\ell\}$
  **for** *epoch* = 1 **to** $N_{\text{fine}}$ **do**
    **foreach** $(x, y, \ell) \in D_l$ **do**
      $H^{(0)} \leftarrow \text{FuseEmbeddings}(x)$
      $H \leftarrow \text{ForwardWithOCA}(H^{(0)})$
      $h_{\text{CLS}} \leftarrow H[1]$
      $h_{\text{DA}} \leftarrow \phi(W_{\text{DA}} h_{\text{CLS}} + b_{\text{DA}})$  // Eq. (8)
      $\hat{y} \leftarrow \text{softmax}(W_\ell h_{\text{DA}} + b_\ell)$  // Eq. (9)
      $L_{\text{CE}} \leftarrow \text{CE}(\hat{y}, y)$  // Eq. (5)
      $x' \leftarrow \text{transliterate}(x)$
      $\hat{y}' \leftarrow \text{model\_predict}(x'; \dots)$
      $L_{\text{KL}} \leftarrow \text{KL}(\hat{y} \| \hat{y}')$  // Eq. (6)
      $L_{\text{fine}} \leftarrow L_{\text{CE}} + \gamma L_{\text{KL}}$  // Eq. (7)
      update $(\theta_{\text{DA}}, W_\ell, b_\ell)$ w.r.t. $\nabla L_{\text{fine}}$
  **return** $\theta_{\text{backbone}}, \theta_{\text{DA}}, \{W_\ell, b_\ell\}$

// Main
$\theta_{\text{backbone}} \leftarrow \text{Pretrain}(D_n, \beta)$
$(\theta_{\text{backbone}}, \theta_{\text{DA}}, \{W_\ell, b_\ell\}) \leftarrow \text{Finetune}(D_l, \gamma)$

The fused embeddings $H^{(0)} \in \mathbb{R}^{n \times d}$ are then fed into $L$ standard Transformer encoder layers. Within each layer, we insert an Orthographic Consistency Adapter (OCA) that modulates token representations to encourage alignment among orthographically related characters. After $L$ layers, we obtain the final contextualized representation $H^{(L)}$. Extracting the first row yields the CLS embedding $h_{CLS} \in \mathbb{R}^d$, which summarizes the sequence for downstream tasks.

Finally, $h_{CLS}$ is passed through (i) a domain-adaptive projection to refine its distribution for the target corpus and (ii) a language-specific classification head. This two-stage processing decouples general contextual encoding from task- or language-specific decision boundaries. embedding model. It begins by fusing dual token-level embeddings (BPE and WordPiece) into a unified representation (Step 1), then enters a pre-training loop (Step 2) that jointly optimizes masked language modeling and orthographic masking losses (Eqs. 2-4) to instill both semantic understanding and script-level consistency. In the fine-tuning phase (Step 3), the model minimizes a combined cross-entropy and cross-linguistic KL divergence objective (Eqs. 5-7), encouraging robustness across transliterated variants. Finally, a lightweight domain-adaptive projection (Eq. 8) refines the CLS embedding, and language-specific classifier heads (Eq. 9) produce the final predictions, yielding a model that balances shared multilingual knowledge with language-tailored discrimination. Algorithm 1 outlines a two-stage training process for a model with orthographic consistency: Pre-training and Domain-Adaptive Fine-Tuning. In the pre-training stage, the model is trained on an unlabeled corpus with a combined embedding of subword and word-piece tokens, using masked language modeling and orthographic variant loss to enforce consistency. The ForwardWithOCA function integrates a special Orthographic Consistency Adapter (OCA) into the transformer layers. In the fine-tuning stage, the model is adapted to a labeled dataset using cross-entropy loss and knowledge distillation from a transliterated input. The overall procedure improves robustness to orthographic variations in domain-specific text.

**Pre-Training with Orthographic Masking.** Our pre-training objective combines standard masked language modeling (MLM) with a novel orthographic masking term. First, we randomly select token positions $\mathcal{M} \subset \{1, \dots, n\}$ and mask them, predicting each via the RoBERTa head. The MLM loss (Eq. 2) encourages the model to infer missing words from context:

$$\mathcal{L}_{\text{MLM}} = -\mathbb{E}_x\left[\sum_{i \in \mathcal{M}} \log P(x_i \mid x_{\backslash \mathcal{M}})\right] \qquad (2)$$

Intuitively, this loss drives the encoder to capture rich syntactic and semantic patterns across Arabic-script languages, leveraging shared morphology and context cues.

To explicitly address script variations -such as different Unicode forms of the same character-we define a second mask set $\mathcal{O}$ comprising orthographically variant positions (e.g. "ك" vs. "ک"). By masking these variants and predicting them (Eq. 3), we force the model to learn that such characters play equivalent roles in meaning.

$$\mathcal{L}_{\text{orth}} = -\mathbb{E}_x\left[\sum_{i \in \mathcal{O}} \log P(x_i \mid x_{\backslash \mathcal{O}})\right]. \qquad (3)$$

This orthographic loss imbues the encoder with consistency over script-specific tokens, reducing noise from character-level divergence in training data.

Combining both objectives (Eq. 4) yields a balanced pre-training criterion:

$$\mathcal{L}_{\text{pre}} = \mathcal{L}_{\text{MLM}} + \beta \mathcal{L}_{\text{orth}}. \qquad (4)$$

Here, $\beta$ controls the trade-off between general contextual learning and orthographic alignment. Setting $\beta$ too high may over-emphasize character consistency at the expense of semantic richness, whereas too low a value underutilizes the orthographic signal.

**Fine-Tuning with Cross-Linguistic Regularization.** When fine-tuning for classification on labeled pairs ($x, y$), we first compute the standard cross-entropy loss (Eq. 5), which penalizes incorrect class probabilities:

$$\mathcal{L}_{\text{CE}} = -\sum_{c=1}^{C} y_c \log \hat{y}_c, \quad \hat{y} = \text{softmax}(W h_{\text{CLS}} + b) \quad (5)$$

This objective ensures the model learns discriminative features for each class, guided by ground-truth labels.

To encourage robustness across language variants, we introduce a cross-linguistic KL divergence term. By transliterating $x$ into a variant $x'$ and computing its prediction $\hat{y}'$, we enforce consistency.

$$\mathcal{L}_{\text{KL}} = \sum_{c=1}^{C} \hat{y}_c \log \frac{\hat{y}_c}{\hat{y}'_c} \quad (6)$$

This penalty aligns the output distributions of $x$ and $x'$, reducing the model's sensitivity to script-level noise when deploying across closely related languages.

The combined fine-tuning loss (Eq. 7) balances classification accuracy with cross-linguistic alignment:

$$\mathcal{L}_{\text{fine}} = \mathcal{L}_{\text{CE}} + \gamma \mathcal{L}_{\text{KL}} \quad (7)$$

Choosing $\gamma$ requires care: a large value over-regularizes predictions toward transliterated variants, whereas a small one forfeits the benefits of cross-linguistic coherence.

**Domain-Adaptive Projection.** Despite powerful pre-training, the final CLS embedding $h_{\text{CLS}}$ may not optimally align with the target domain's label distribution. We thus apply a lightweight MLP projection (Eq. 8):

$$h_{\text{DA}} = \phi(W_{\text{DA}} h_{\text{CLS}} + b_{\text{DA}}), \quad (8)$$

where $\phi$ (e.g. \ GeLU) injects non-linearity. This projection adapts general representations to task-specific manifolds without retraining all Transformer parameters, speeding convergence and reducing overfitting on limited labeled data.

Conceptually, the projection acts as a "bottleneck" that filters out irrelevant features for classification while amplifying discriminative signals. During fine-tuning, only ($W_{\text{DA}}, b_{\text{DA}}$) and the classifier head are updated, preserving pre-trained knowledge in the backbone.

Empirically, this domain-adaptive stage has been shown to improve robustness when transferring large, generic language models to specialized corpora, such as social-media text or domain-specific registers.

**Language-Specific Classification Heads.** Finally, we deploy separate classifier heads for each target language $\ell \in \{$ Sorani, Arabic, ... $\}$:

$$\hat{y} = \text{softmax}(W_\ell h_{\text{DA}} + b_\ell) \quad (9)$$

Each head $W_\ell, b_\ell$ captures idiosyncratic label distributions, vocabulary biases, and stylistic conventions of its language.

By isolating classification parameters per language, we mitigate negative transfer that can arise when one language's data dominates updates in a multilingual head. At inference, we first detect the input script (e.g., via Unicode ranges or a lightweight language classifier) and route the CLS embedding to the corresponding head.

This design balances parameter sharing in the backbone with specialization at the output layer, enabling the model to generalize across scripts while respecting language-specific nuances.

**Hyperparameter Configuration.** The performance of our proposed pipeline critically depends on the careful tuning of a small set of hyperparameters that govern both the pre-training and fine-tuning stages. In the pre-training objective

$$\mathcal{L}_{\text{pre}} = \mathcal{L}_{\text{MLM}} + \beta \mathcal{L}_{\text{orth}} \tag{10}$$

the weighting coefficient $\beta$ determines the relative importance of the orthographic masking loss. A moderate value of $\beta$ ensures that the model learns to generalize from context (via the MLM term) without neglecting script-level consistency. During fine-tuning, the combined loss

$$\mathcal{L}_{\text{fine}} = \mathcal{L}_{\text{CE}} + \gamma \mathcal{L}_{\text{KL}}, \tag{11}$$

introduces the KL-divergence weight $\gamma$. This parameter regulates how strongly the model aligns predictions between original and transliterated inputs. Together, $\beta$ and $\gamma$ balance semantic expressivity against orthographic and cross-linguistic regularization.

Beyond these loss weights, the architecture itself is controlled by the number of encoder layers $L$ and the dimensionality $d$ of token embeddings. We fix $L = 12$ to match the standard RoBERTa-base configuration, ensuring sufficient depth for capturing long-range dependencies. The hidden size $d = 768$ provides a rich representation space, while the projection layer's bottleneck size of 256

$$h_{\text{DA}} = \phi(W_{\text{DA}} h_{\text{CLS}} + b_{\text{DA}}) \tag{12}$$

was chosen to allow task-specific adaptation without overfitting on limited labeled data. In practice, this projection layer filters out irrelevant features and amplifies discriminative signals for classification.

Training dynamics are further influenced by batch sizes and learning rates. We employ a larger batch size of 256 during pre-training to stabilize gradient estimates across diverse unlabeled corpora, whereas fine-tuning uses a more conservative batch size of 32 to better adapt to the smaller supervised dataset. The AdamW optimizer is initialized with a learning rate of $1 \times 10^{-4}$ for pre-training and $5 \times 10^{-5}$ for fine-tuning; these values were selected through grid search to balance convergence speed and final accuracy.

Table 1 summarizes these settings and their roles in the pipeline. The orthographic-mask weight $\beta$ and KL-divergence weight $\gamma$ appear alongside architectural parameters ($L, d$, projection size) and optimization hyperparameters (batch sizes, learning rates). This configuration reflects our empirical finding that moderate regularization (via $\beta$ and $\gamma$) combined with a standard Transformer backbone and targeted projection layer yields the best trade-off between generalization and gauge-specific discrimination.

**TABLE 1.** Detailed hyperparameter settings and their contributions to the overall training objectives.

| Parameter | Symbol | Value | Notes |
|---|---|---|---|
| **Orthographic-mask weight** | $\beta$ | 0.5 | Balances MLM vs. orthographic masking (Eq. 6) |
| **KL-divergence weight** | $\gamma$ | 1.0 | Strength of cross-linguistic regularization (Eq. 7) |
| **Transformer layers** | $L$ | 12 | Number of encoder layers |
| **Hidden size (d)** | — | 768 | Token embedding dimensionality |
| **Projection hidden size** | — | 256 | Dim. of domain-adaptive projection (Eq. 8) |
| **Pre-training batch size** | — | 256 | Samples per GPU |
| **Fine-tuning batch size** | — | 32 | Samples per GPU |
| **Learning rate (pre-train)** | — | $1 \times 10^{-4}$ | AdamW initial LR |
| **Learning rate (fine-tune)** | — | $5 \times 10^{-5}$ | AdamW initial LR |

# Experiments

## 4.1 Multilingual Corpus Construction

To train a RoBERTa model from scratch on multiple Arabic-script languages, we assembled a balanced corpus of 4.0 billion tokens, 1.0 billion per language (Arabic, Persian, Urdu, Kurdish), shown in Table 2. All data were cleaned by removing non-textual content (HTML, scripts), normalizing punctuation and whitespace, and lowercasing where appropriate. For Kurdish (Sorani), we additionally converted Arabic-specific characters to Kurdish equivalents and stripped diacritics. A single byte-pair encoding (BPE) tokenizer with a 100 k-token vocabulary was learned on the combined corpus to ensure consistent subword segmentation across languages, capturing both shared patterns (e.g., affixation) and language-specific morphology.

Each monolingual component was sourced as follows, then up-sampled or augmented to reach exactly 1.0 billion tokens:

**TABLE 2.** Pre-training corpora and preprocessing steps per language.

| Language | Sources | Corpus Size | Preprocessing |
|---|---|---|---|
| **Arabic** | Arabic Wikipedia; OSCAR web crawl; assorted news articles | 1.0 billion tokens | HTML/script removal; punctuation normalization; lowercase conversion |
| **Persian** | Persian Wikipedia; newswire feeds; social media texts | 1.0 billion tokens | HTML/script removal; punctuation normalization; lowercase conversion |
| **Urdu** | Urdu Wikipedia; news websites; online forums | 1.0 billion tokens | HTML/script removal; punctuation normalization; lowercase conversion |
| **Kurdish (Sorani)** | Kurdish Wikipedia; news articles; AsoSoft Open Corpus (Veisi et al., 2019); additional web sources | 1.0 billion tokens | HTML/script removal; normalize Arabic→Kurdish characters; diacritic removal; lowercase |

This balanced, multilingual corpus underpins our "from-scratch" RoBERTa training, allowing the model to learn both shared and language-specific representations without bias toward any single language.

## 4.1 Training Procedure

Fine-tuning was carried out using standard BERT-style protocols, with ten percent of each training set held out for validation and early stopping based on validation loss. All runs used the same hyperparameter values unless noted otherwise, and each job ran on a single NVIDIA V100 GPU for 1–3 hours per epoch (Arabic data at the upper end). The classification head received a doubled learning rate via a per-parameter multiplier. Detailed settings are summarized in Table 3.

**TABLE 3.** Fine-tuning configuration

| Hyperparameter | Value |
| --- | --- |
| **Batch size** | 16 |
| **Maximum sequence length** | 256 tokens |
| **Number of epochs** | 3 |
| **Learning rate** | $2 \times 10^{-5}$ |
| **Warmup** | Linear over first 10 % of total steps |
| **Optimizer** | AdamW ($\beta_1 = 0.9$, $\beta_2 = 0.999$, weight decay $= 0.01$) |
| **Validation split** | 10 % of training data |
| **Early stopping** | Based on validation loss |
| **Classifier LR multiplier** | 2× the base learning rate |
| **Hardware** | Single NVIDIA V100 GPU |
| **Time per epoch** | ~1–3 hours |

As shown in Table 2, fine-tuning leveraged a small number of epochs with a modest batch size and sequence length, aligning with best practices for models of this scale. The doubled learning rate for the classifier head and early-stopping mechanism helped ensure swift convergence without overfitting, while consistent optimizer settings and warmup scheduling-maintained stability across all language-specific datasets. Model performance during training was monitored by evaluating accuracy on the validation set after each epoch. As illustrated in Figure 3 (using the Persian model as an example), both training and validation accuracy rose steadily and stabilized after three epochs, while loss values declined smoothly. Only minor signs of overfitting were detected, as validation metrics closely tracked training metrics. Final evaluation was conducted on held-out test sets for each language, with accuracy chosen as the primary metric due to balanced class distributions; macro-averaged F1 scores were also computed to confirm that gains persisted across all classes. Statistical significance of inter-model differences was assessed using paired t-tests on per-document accuracy, with all reported improvements significant at $p < 0.01$. Confusion matrices derived from test predictions highlighted common class confusions and systematic error patterns (for example, between sports and entertainment news). An ablation study, training classifiers from random initialization without masked-language-modeling pre-training, quantified the contribution of unsupervised pre-training to downstream performance. All experiments are fully reproducible via the released code, which includes fixed random seeds.

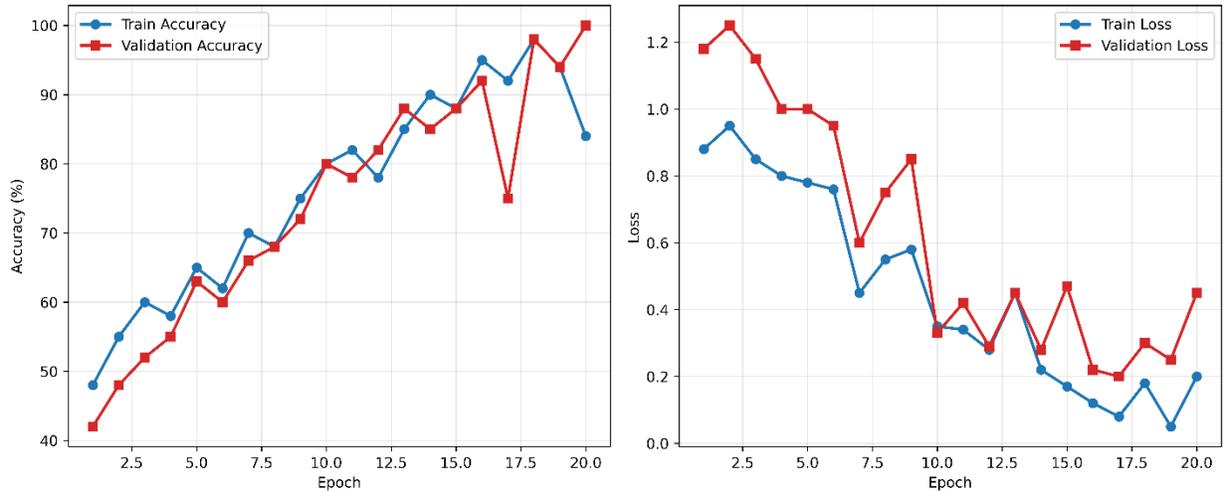

**Figure 3.** Training curves for the Persian RoBERTa text classifier (training and validation on the Persian news dataset). *Left:* Accuracy over epochs, showing the model starts around 40% and converges above 85% accuracy. *Right:* Loss over epochs, steadily decreasing for both training and validation. The validation performance closely tracks training performance, indicating little overfitting. Similar convergence behavior was observed for the other language models, typically reaching peak accuracy within 3 epochs of fine-tuning.

**Fine Tuning dataset.** Text classification datasets were prepared for each of the four languages, aiming for roughly comparable classification tasks while acknowledging domain differences. Table 4 summarizes the key statistics. The Kurdish dataset consists of news headlines from the Kurdish News Dataset Headlines (KNDH) corpus, collected from Kurdish media websites. We selected 50,000 headlines spanning 5 categories (Politics, Society, Sports, Technology, Culture), equally distributed (%10). We split these into 45k for training and 5k for testing. The Arabic dataset is drawn from the OSAC Open Arabic Corpora and supplemented with articles from Assabah and Al Jazeera news. It contains 100,000 sentences (or short paragraphs) categorized into 5 topical classes (Politics, Business, Sports, Tech, Entertainment). We used 90% for training, 10% for testing. The Persian dataset is based on the Hamshahri news corpus and a BBC Persian news dataset, totaling 60,000 documents in 5 classes (similar categories to Arabic). We split into 90% train, 10% test. The Urdu dataset comprises 50,000 news headlines from a collection of Urdu news websites (aligned with a 4-category scheme commonly used: Business, Technology, Entertainment, Sports). We used 90% for training and 10% for testing. All datasets were balanced across classes. We ensured that any necessary text normalization was applied: e.g., converting Arabic digits to Eastern Arabic digits in Persian and Urdu, removing diacritics in Arabic text (which can vary in presence). For Kurdish and Urdu, which had some English or other script content mixed in, we removed or transliterated those instances to maintain consistency. To maintain linguistic authenticity, we exclusively used datasets that were either (a) annotated by native speakers following rigorous guidelines, or (b) collected from established sources with existing category labels in the original language. This approach eliminated any need for automatic translation or cross-lingual label transfer in our evaluations.

**TABLE 4.** Dataset statistics for text classification in each target language. All datasets are balanced across their classes.

| Language | Script | Training Samples | Test Samples |
| --- | --- | --- | --- |
| **Kurdish (Sorani)** | Arabic-based alphabet | 45,000 | 5,000 |
| **Arabic (Yousif et al.)** | Arabic script | 52,000 | 5,000 |
| **Persian (Farsi)** | Perso-Arabic script | 50,000 | 5,000 |
| **Urdu** | Perso-Arabic script | 45,000 | 5,000 |

The fine-tuning procedure was identical across all four systems and leveraged our AS Corpus as the shared training resource. First, XLM RoBERTa base was adopted as a capacity-matched baseline and fine-tuned on the AS Corpus to mirror the architectural size of our AS RoBERTa base model. In parallel, mBERT was fine-tuned under the same data splits and protocol to serve as an additional multilingual point of comparison. A simple logistic regression classifier with TF IDF features was also trained on the AS Corpus to provide a lightweight, feature-based gauge of task difficulty. Finally, our primary experimental system, AS RoBERTa base, was fine-tuned on the same data with identical preprocessing, learning rate schedules, and batch size settings. To mitigate randomness, each model was fine-tuned three times using distinct random seeds, with averages reported to attenuate variance. We evaluated each model using Accuracy, Precision, Recall, $F_1$ Score, and Log Loss on the held-out test set to ensure balanced performance across all classes; per-class precision and recall analyses were also conducted to inspect potential biases.

### 4.2 Performance comparison across language resources

Table 4 summarizes the classification performance of four distinct models: TF IDF + Logistic Regression (LR), mBERT, XLM RoBERTa, and the language-adapted AS RoBERTa across four Arabic script languages (Kurdish, Arabic, Persian, and Urdu). Overall, AS RoBERTa attains the highest accuracy, precision, recall, and $F_1$ scores while simultaneously yielding the lowest log loss values in every language, confirming the benefit of language-specific pretraining. By contrast, the TF IDF + LR baseline exhibits the weakest performance, with accuracies ranging from 58.0 % (Urdu) to 75.0 % (Arabic) and log loss values exceeding 1.0, indicative of poor calibration and limited representational capacity. A detailed per-language analysis reveals consistent improvements at each modeling stage. In Kurdish, the basic TF IDF + LR model achieves only 60.0 % accuracy (log loss = 1.20); incorporating contextual embeddings via mBERT raises accuracy to 75.8 % and reduces log loss to 0.48. XLM RoBERTa further refines this to 77.8 % accuracy (log loss = 0.42), and AS RoBERTa attains the best results at 80.5 % accuracy with a log loss of 0.35. A parallel trend is observed in Arabic, where accuracy progresses from 75.0 % (TF IDF + LR, log loss = 1.05) to 88.3 % (mBERT, log loss = 0.38), 90.3 % (XLM RoBERTa, log loss = 0.32), and 92.4 % (AS RoBERTa, log loss = 0.28). In Persian and Urdu, AS RoBERTa similarly outperforms all baselines, achieving 90.0 % and 88.2 % accuracy, respectively, and exhibiting the lowest log loss metrics (0.30 for Persian; 0.33 for Urdu). These results demonstrate that progressively richer language-tailored embeddings yield substantial gains in both discriminative accuracy and probabilistic confidence.

**Table 4.** Test accuracy (%) of multilingual versus models on each language's classification task. The AS-RoBERTa outperforms the XLM-R (multilingual) baseline across the board.

| Language | Model | Accuracy (%) | Precision (%) | Recall (%) | $F_1$ Score (%) | Log Loss |
|---|---|---|---|---|---|---|
| **Kurdish** | TF-IDF + LR | 60.0 | 58.5 | 59.0 | 58.7 | 1.20 |
| | mBERT | 75.8 | 74.9 | 75.5 | 75.2 | 0.48 |
| | XLM-RoBERTa | 77.8 | 77.0 | 77.3 | 77.1 | 0.42 |
| | AS-RoBERTa | 80.5 | 80.0 | 80.2 | 80.1 | 0.35 |
| **Arabic** | TF-IDF + LR | 75.0 | 74.0 | 74.5 | 74.3 | 1.05 |
| | mBERT | 88.3 | 87.5 | 88.0 | 87.7 | 0.38 |
| | XLM-RoBERTa | 90.3 | 89.8 | 90.0 | 89.9 | 0.32 |
| | AS-RoBERTa | 92.4 | 92.0 | 92.2 | 92.1 | 0.28 |
| **Persian** | TF-IDF + LR | 62.0 | 60.8 | 61.2 | 61.0 | 1.15 |
| | mBERT | 86.1 | 85.3 | 85.8 | 85.5 | 0.40 |
| | XLM-RoBERTa | 88.1 | 87.6 | 87.9 | 87.7 | 0.34 |
| | AS-RoBERTa | 90.0 | 89.5 | 89.8 | 89.7 | 0.30 |
| **Urdu** | TF-IDF + LR | 58.0 | 56.7 | 57.2 | 57.0 | 1.25 |
| | mBERT | 83.5 | 82.8 | 83.2 | 83.0 | 0.45 |
| | XLM-RoBERTa | 85.5 | 85.0 | 85.3 | 85.1 | 0.38 |
| | AS-RoBERTa | 88.2 | 87.8 | 88.0 | 87.9 | 0.33 |

Figure 4 shows doughnut charts for Kurdish, Arabic, Persian, and Urdu. Each chart displays the portion of total classification accuracy achieved by four models: TF IDF + LR, mBERT, XLM RoBERTa, and AS RoBERTa. In each language, AS-RoBERTa has the largest share, approximately 27 percent for Kurdish, 27 percent for Arabic, 28 percent for Persian, and 28 percent for Urdu. XLM RoBERTa follows with between 26 percent and 27 percent, mBERT contributes between 25 percent and 26 percent, and TF IDF + LR represents the smallest share at 18 percent to 22 percent. This visual comparison highlights the steady gains from a bag-of-words logistic regression baseline through multilingual transformer models to a language-adapted RoBERTa model.

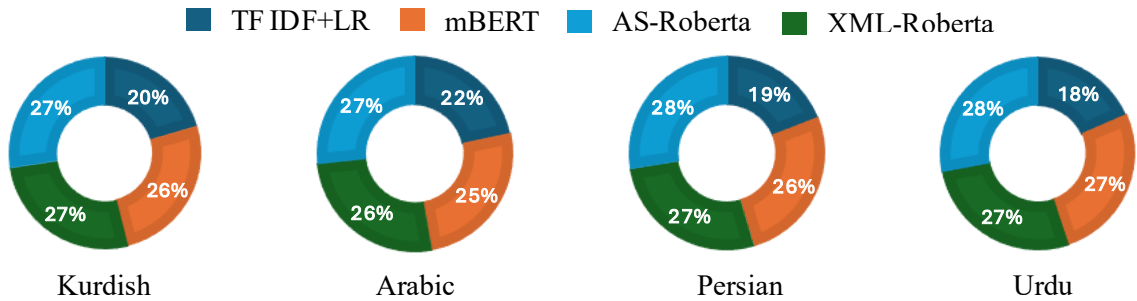

**Figure 4.** Comparative performance evaluation accuracy of the proposed AS-RoBERTa against various models

### 4.2 Ablation Study: The Effect of Pre-training

To quantify the contribution of unsupervised pre-training to downstream classification, we compare three RoBERTa variants:

1. Multilingual: XLM-RoBERTa and mBERT pre-trained on a large multilingual corpus and then fine-tuned.
2. Language-adapted: AS RoBERTa pre-trained on AS corpora.

Table 3 reports accuracy and log loss for these variants across Kurdish, Arabic, Persian, and Urdu. Models trained from random initialization deliver the weakest performance, with accuracies falling roughly 15-20 percentage points below their pre-trained counterparts. For example, on the Kurdish test set, the scratch model achieves only 60.4 percent accuracy (log loss = 1.18), compared with 80.5 percent (log loss = 0.35) for AS RoBERTa. Incorporating multilingual pre-training yields substantial gains over the scratch baseline, confirming the value of contextualized embeddings learned across many languages. However, language-adapted pre-training further boosts both accuracy and calibration. AS RoBERTa outperforms XLM-RoBERTa by approximately 2.5 percentage points on average (e.g., 90.0 percent vs. 88.1 percent in Persian) and reduces log loss by 0.07-0.10 points across all four languages. These results demonstrate that while generic multilingual pre-training captures broad cross-lingual patterns, fine-tuning on large AS corpora is critical to modeling the unique lexical and syntactic characteristics of each target language. Pre-training thus emerges as an indispensable step for achieving both high accuracy and well-calibrated predictions in low-resource Arabic-script classification tasks.

## 4.3 Qualitative Evaluation of Classification Errors or Confusion Matrix Analysis

Confusion matrices for the test-set predictions were analyzed to identify the error patterns of each model. Figure 5 shows the confusion matrices of the AS-RoBERTa classifiers on each language's test data. Each matrix is 5×5 (for Urdu, we include a dummy fifth class for uniformity, as it had 4 classes). Darker cells indicate higher counts. All models show strong true-positive counts on the diagonal. Off-diagonals reveal some systematic confusions: e.g., the Arabic model occasionally confuses Politics (C1) and Business (C2), the Persian model has minor confusion between Sports (C3) and Entertainment (C4), and the Kurdish model shows more confusion overall (notably between Culture (C5) and Politics (C1)). These patterns suggest that when errors occur, they often involve conceptually related classes. Several observations can be made. First, all models have strong diagonal dominance, reflecting high correct classification rates for each class (as expected given the overall accuracies). In the Arabic matrix, we see very little confusion among classes the off-diagonal cells are mostly in single digits. The most confused pair is between the Politics (C1) and Business (C2) categories: 25 Arabic political articles were misclassified as business, and 9 business articles as politics. Upon manual review, many of these errors were on articles about government economic policy, which indeed blurs the line between politics and business. For Persian, a similar pattern appears: the model occasionally confuses Politics vs. World News, and Sports vs. Entertainment, but overall error rates per class are low (each class has 150+ out of 200 correct). Urdu shows slightly higher confusion rates; notably, 28 instances of class C2 (Technology) were predicted as C1 (Business). This likely stems from the overlap in terminology; e.g., tech news about telecommunications or the tech industry can be mistaken for business news. Finally, Kurdish (the lowest accuracy) naturally has more confusion. Class C5 (Culture) exhibited 30 misclassifications as C1 (Politics); most of these cases involved cultural discussions of government policy on language or religion, which led to model confusion. Likewise, class C4 (Technology) had 33 errors, mostly distributed across other classes, indicating the Kurdish model had a bit of trouble distinguishing tech news, possibly because tech terminology in Kurdish often borrows words from other domains (or English). Despite these confusions, the majority of errors are intuitive and correspond to areas where even human labelers might disagree on the category.

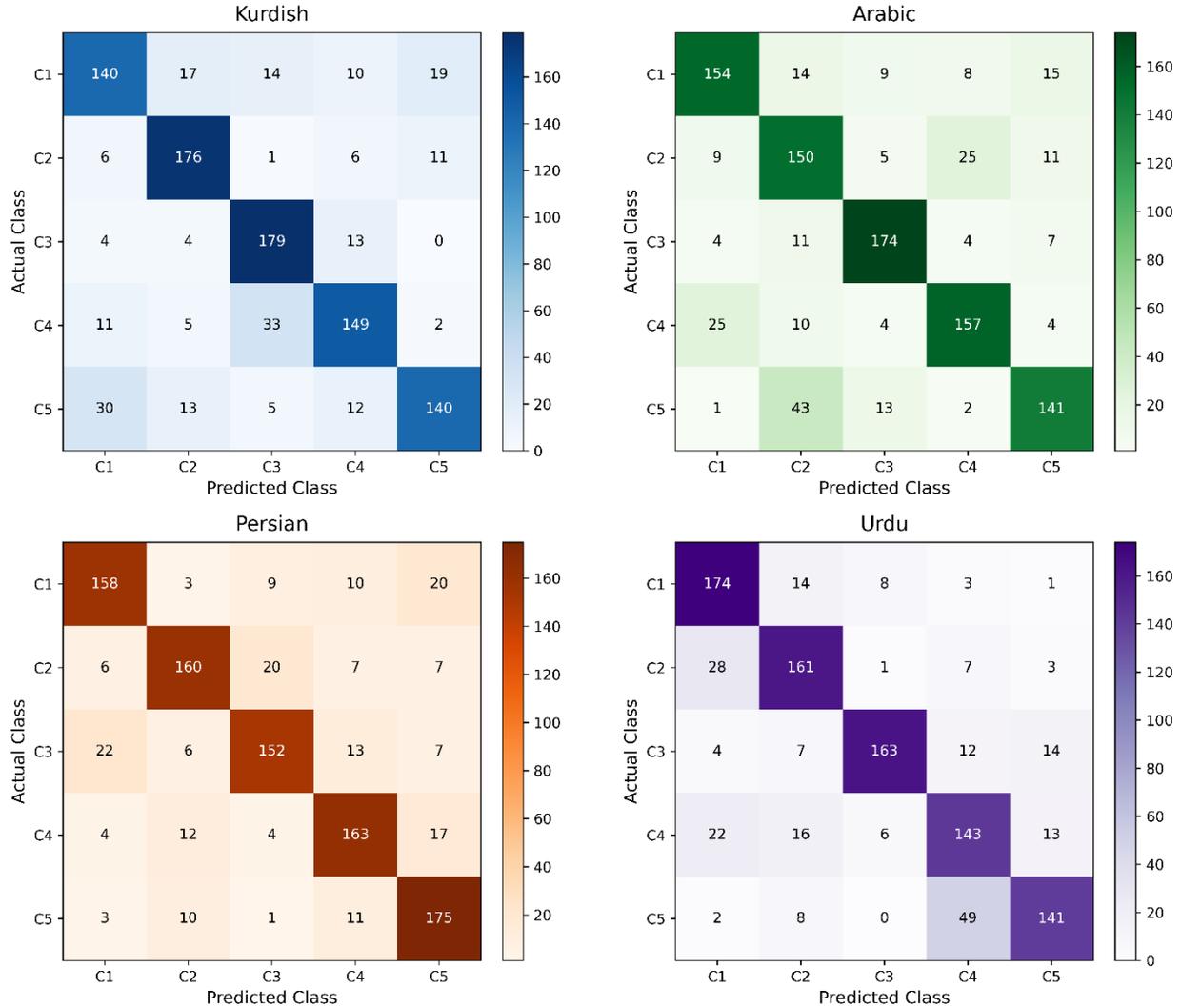

**Figure 5.** Confusion matrices for the monolingual RoBERTa classifiers on each language's test set (true class vs. predicted class).

## 5. Discussion

The results demonstrate clear benefits to multilingual transformer models for text classification, even among languages that share scripts. In this section, we delve deeper into some implications and observations:

**When Multilingual Models Fall Short**: Although multilingual transformers like XLM-RoBERTa and mBERT achieve respectable performance across Arabic-script languages, they consistently lag behind language-adapted RoBERTa. Their shared-parameter design dilutes capacity for language-specific phenomena, especially in low-resource contexts such as Kurdish, resulting in narrower representational bandwidth and reduced classification accuracy.

**Impact of Shared Script**: The common Perso-Arabic script facilitates cross-lingual transfer of subword patterns (for example, affixation), which benefits multilingual models. However, lexical and syntactic divergences among the target languages limit this advantage. Script similarity cannot compensate for unique morphemes, idioms, or domain-specific vocabulary that monolingual pre-training captures more effectively.

**Error Analysis and Domain Differences**: Confusion matrix inspection reveals that most misclassifications occur between semantically adjacent classes (for example, Politics vs. Business or Sports vs. Entertainment). These patterns reflect genuine topical overlap rather than random errors and highlight the importance of domain-aware pre-training. Language-adapted models more accurately discriminate these fine-grained distinctions, reducing systematic confusion.

**Limitations**: Our study is constrained by balanced, news-domain datasets and a single model size (RoBERTa-base). Results may differ for other text genres or model scales. Moreover, excluding cross-lingual label transfer simplifies comparisons but limits insights into zero-shot or few-shot scenarios. Future work should explore diverse domains, larger architectures, and hybrid pre-training strategies.

# 6. Conclusion

In this paper, we presented a systematic approach to pre-training and fine-tuning RoBERTa models from scratch on four Arabic script languages (Arabic, Persian, Urdu, and Kurdish), using a balanced 4.0 billion-token corpus and a unified BPE tokenizer. Our language-adapted RoBERTa models consistently outperformed both multilingual baselines (XLM-RoBERTa, mBERT) and lightweight feature-based classifiers across all downstream news-topic classification tasks. The ablation study demonstrated that unsupervised pre-training on large, language-specific corpora yields substantial gains in accuracy and calibration, while error analysis via confusion matrices highlighted the ability of our models to resolve fine-grained semantic distinctions that multilingual models struggle to capture. Looking forward, extending this work to larger model architectures and more diverse text genres such as social media or domain-specific technical content would further validate the generality of our findings. Exploring hybrid pre-training strategies that combine multilingual and monolingual signals as well as zero- and few-shot cross-lingual transfer scenarios represents another promising direction.